\documentclass[final,3p,times]{elsarticle}

\usepackage{mathtools}

\usepackage{amssymb}
\usepackage{subcaption}

\usepackage{amsthm}
\usepackage{amsmath}
\usepackage{enumitem}
\usepackage{booktabs}
\usepackage{multirow}
\usepackage{lscape}
\usepackage{tabu}
\usepackage{threeparttable}
\usepackage{natbib}

\usepackage{tikz}
\usetikzlibrary{shapes.geometric, arrows}
\tikzstyle{startstop} = [rectangle, rounded corners, minimum width=3cm, minimum height=1cm,text centered, draw=black, fill=red!30]
\tikzstyle{input} = [trapezium, trapezium left angle=60,trapezium right angle=120,minimum width=1cm, minimum height=1cm, text centered, draw=black, fill=blue!5, text width=2cm]
\tikzstyle{output} = [trapezium, trapezium left angle=120,trapezium right angle=60, minimum width=1cm, minimum height=1cm, text centered, draw=black, fill=blue!20, text width=3cm]
\tikzstyle{NN} = [trapezium, minimum width=4.5cm, minimum height=1cm, text centered, draw=black, fill=gray!40]

\tikzstyle{process} = [rectangle, rounded corners, minimum width=4.5cm, minimum height=1cm, text centered, draw=black, fill=gray!20]
\tikzstyle{decision} = [diamond, minimum width=3cm, minimum height=1cm, text centered, draw=black, fill=green!5]
\tikzstyle{data} = [rectangle, rounded corners, minimum width=1cm, minimum height=1cm,text centered, draw=black, fill=red!30]
\tikzstyle{arrow} = [thick,->,>=stealth]
\tikzstyle{arrow2} = [dashed,->,>=stealth]

\usepackage[pagewise]{lineno}

\usepackage{setspace}

\journal{Accident Analysis \& Prevention Journal}

\begin{document}

\begin{doublespacing}
\begin{frontmatter}
    
    
    
    \title{\textbf{Crash Data Augmentation Using Conditional Generative Adversarial Networks (CGAN) for Improving Safety Performance Functions}}

    \author{Mohammad Zarei $^1$}
    \author{Bruce Hellinga $^2$}

    \address{$^1$ Ph.D. Candidate, Department of Civil and Environmental Engineering, University of Waterloo, 200 University Ave., Waterloo, ON N2L3G1, Canada (corresponding author). E-mail: mzarei@uwaterloo.ca}
    \address{$^2$ Professor, Department of Civil and Environmental Engineering, University of Waterloo, 200 University Ave., Waterloo, ON N2L3G1, Canada. E-mail: bruce.hellinga@uwaterloo.ca }

    \begin{abstract}
        In this paper, we present a crash frequency data augmentation method based on Conditional Generative Adversarial Networks to improve crash frequency models. The proposed method is evaluated by comparing the performance of Base SPFs (developed using original data) and Augmented SPFs (developed using original data plus synthesised data) in terms of hotspot identification performance, model prediction accuracy, and dispersion parameter estimation accuracy. The experiments are conducted using simulated and real-world crash data sets. The results indicate that the synthesised crash data by CGAN have the same distribution as the original data and the Augmented SPFs outperforms Base SPFs in almost all aspects especially when the dispersion parameter is low.

    \end{abstract}
    
    \begin{keyword}
       Conditional Generative Adversarial Networks (CGAN) \sep Hotspot identification \sep Crash Data Augmentation \sep Safety performance function \sep Crash data simulation
    \end{keyword}

\end{frontmatter}
\end{doublespacing}


\section{Introduction}
\label{section:intro}

Safety Performance Functions (SPFs) have been widely used by researchers and practitioners to conduct roadway safety evaluation. They are crash frequency prediction models that are calibrated to crash data using either parametric models such as negative binomial model \cite{lord2010crashdata} or non-parametric models such as neural networks \cite{pan2017NN, zarei2021cgan, zarei2021real}. 

One of the main challenges in developing SPFs is to cope with small crash data sets and limited observed crashes (i.e. preponderance of zeros) \cite{lord2010crashdata}. The desirable large-sample properties of various parameter-estimation approaches (for example, maximum likelihood estimation) are not achieved with small sample sizes. There are several approaches to address this issue and increase the size of a crash data set. A common method is to use the crash data sets for a number of consecutive years (e.g., 3-5 years) \cite{HSM}. However, it has to be assumed that road network and the related parameters did not change in that period and the exposure data have to be estimated for all years \cite{ambros2016developing}. Furthermore, sometimes the network contains so few sites of a certain type (e.g. roundabouts, ramps) that reliable SPFs cannot be developed even when expanding the number of years of crash data. In this case, the common practice is to combine them with other site types and include categorical variables within the SPF to distinguish site type or to use an SPF developed using data from another jurisdiction. 

With the advance of deep learning models, a powerful method to deal with imbalance or small size data sets in the literature is using generative models such as conditional generative adversarial network (CGAN). CGAN has been successfully used in the literature for preparing balanced data sets which is critical for developing proper classification models \cite{islam2021crash,cai2020real, frid2018synthetic}. However, such oversampling methods have not been used for improving SPF development as a regression model. In this paper, we propose a data augmentation method based on CGAN and evaluate its performance using both simulated and real-world crash data sets. 

The remainder of this paper is organized as follows. The next section presents a background on SPFs, crash data augmentation, and CGAN. Section \ref{section:Methodology} describes the proposed crash data augmentation methodology. The evaluation process and results are discussed in Section \ref{section:eval}. Finally, conclusions and recommendations for future works are presented in section \ref{section:conclusion}.

\section{Background and related works}
SPFs are regression models for estimating the average crash frequency of road segments or intersections. Traditionally, SPFs are developed using a statistical model such as negative binomial (NB) model including both traffic and geometric factors. An example of such a parametric SPF based on NB model might be as follows in terms of generic functional form \cite{hauer2015art}:

\begin{equation}
    \label{eq:spf}
    \mu = \exp(\beta_0 + \beta \times \ln(\text{AADT}) + \beta_1 X_1 + \beta_2 X2 + ... + \beta_n X_n)
\end{equation}

\noindent where $\mu$ is the predicted crash frequency, AADT is annual average daily traffic, $X_1$, $X_2$, … , $X_n$ are $n$ roadway geometric variables, and $\beta_0$, $\beta$, $\beta_1$, $\beta_2$, $\beta_n$  are the regression coefficients. 

One of the main applications of SPFs is for crash hotspot identification which is also referred to as network screening. Conventional network screening uses an empirical Bayes (EB) approach first proposed by Hauer \cite{hauer1997observational} in which the EB estimate for long-term crash mean is calculated using a weighted average between the predicted crash frequency from SPFs and the observed number of crashes. Then sites will be ranked based on their EB estimates to find top hotspots for further investigation and safety improvement \cite{HSM}. As a result, the accuracy of SPFs can have significant impacts on network screening results. 
The SPF development process can be quite challenging due to methodological issues associated with crash data such as low sample mean, over-dispersion, time-varying explanatory variables, temporal and spatial correlation, and low sample size  to name but a few \cite{lord2010crashdata}. In this paper, we aim to propose a data augmentation method for improving SPFs by addressing the low sample size issue.

To the best knowledge of the authors, crash data augmentation has not been used for SPF development. However, there are several works that used different methods to augment (or balance) crash data sets for classification purposes (e.g. real time crash prediction, crash severity prediction). In general, there are three main data balancing methodologies used in the literature to handle imbalanced crash datasets including random under-sampling of majority class \cite{yu2018impact,xu2012evaluation}, random over-sampling of minority class \cite{parsa2019real,li2020application}, and using deep generative models \cite{islam2021crash, cai2020real, chen2021novel}. Deep generative models are neural network based models which are trained to represent an estimation of the underlying distribution of data \cite{goodfellow2016nips} and have been shown to perform better than the other two mentioned methods for data augmentation purposes \cite{islam2021crash, cai2020real,chen2021novel}.  A powerful type of deep generative models is deep generative adversarial networks (GANs) that has been widely used in various fields \cite{gui2020review}. In crash data modeling for instance, recent studies \cite{cai2020real, chen2021novel} have shown that real time crash prediction models based on augmented data sets generated by GAN provided the best prediction accuracy as GAN is able to generate data that more closely mimics the characteristics of the real data.

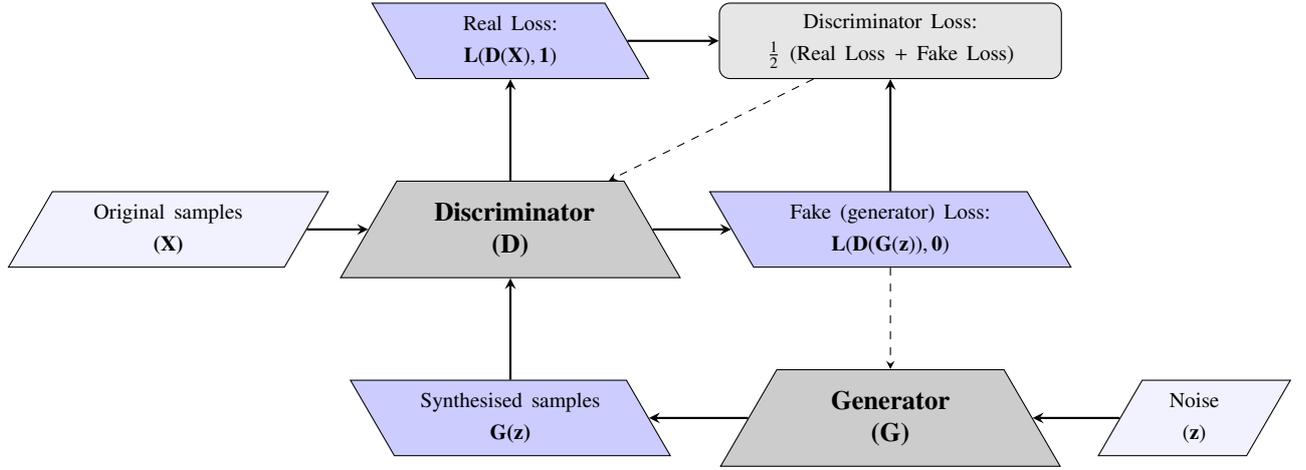
\begin{figure}[tbp]
    \centering
    
    \begin{tikzpicture}[auto, node distance=4cm,>=latex']
        \node (D) [NN, text width = 2cm] {\textbf{Discriminator (D)}};
         \node (G_out) [output, below of = D, yshift = 1.5cm, text width = 2.5cm] {\footnotesize Synthesised samples \\ \textbf{G(z)}};
        
        \node (G) [NN, right of = G_out, xshift = 1cm, text width = 2cm] {\textbf{Generator (G)}};
        
        \node (noise) [input, right of = G,yshift = 0cm, text width = 1cm] {\footnotesize Noise (\textbf{z})};

        \node (real) [input, left of=D, text width = 1.5cm, xshift = -0.5cm, text width = 2.5cm] {\footnotesize Original samples \\\textbf{(X)}};
        
        \node (D_outF) [output, right of= D, xshift = 1cm, text width = 3cm] {\footnotesize Fake (generator) Loss:\\ $\mathbf{L(D(G(z))}, \mathbf{0})$};
        \node (D_outR) [output, above of= D, yshift = -1.5cm,xshift = 0 cm, text width = 2cm] {\footnotesize Real Loss: $\mathbf{L(D(X)}, \mathbf{1})$};
        
        \node (totL) [process, above of= D_outF, yshift = -1.5cm,xshift = 0 cm, text width = 4cm] {\footnotesize Discriminator Loss: \\ $\frac{1}{2}$ (Real Loss + Fake Loss)};
        
        \draw [arrow] (noise) -- (G);
        \draw [arrow] (G) -- (G_out);
        \draw [arrow] (G_out) -- (D);
        \draw [arrow] (real) -- (D);
        \draw [arrow] (D) -- (D_outF);
        \draw [arrow] (D) -- (D_outR);
        \draw [arrow] (D_outR) -- (totL);
        \draw [arrow] (D_outF) -- (totL);
        \draw [arrow2] (totL) -- (D);
        \draw [arrow2] (D_outF) -- (G);
 
    \end{tikzpicture}
    
    \caption{GAN training structure}
    \label{fig:GAN}
\end{figure}

GAN includes competitive training of two neural networks (generator G and discriminator D in Figure \ref{fig:GAN}). The generator's goal is to generate samples from the same distribution as the training data, and the discriminator's goal is to determine if the data are real or fake \cite{goodfellow2020generative}. Since both networks are trained simultaneously based on each other feedback, both networks are forced to increase their performances after each cycle. This process can be formalized as a min-max function:

\begin{equation}
     \underset{G}{\min} \, \underset{D}{\max} \, V(D,G) = \mathbb{E}_{x \sim p(x)} \left[\log(D(x)) \right] + \mathbb{E}_{z \sim p(z)} \left[\log\left(1-D(G(z))\right) \right]
\end{equation}

\noindent where $p(x)$ is the training data distribution, $p(z)$ is the prior distribution of the generative network, and $z$ is a noise vector sampled from the model distribution $p(z)$ such as the Gaussian or uniform distribution. 

In Figure \ref{fig:GAN}, $D(X) \in [0,1]$ is the output of the discriminator (real/fake classifier) using real data $(X)$ as input, and $G(z)$ is the output of generator using noise $(z)$ as input. Real loss value shows the ability of the discriminator to recognize the real instances (i.e. $X$), and is calculated based on $D(X)$, unit vector (i.e. $\mathbf{1}$) and a loss function (e.g. binary cross entropy). Fake loss shows the ability of the discriminator to recognize fake instances (i.e. $G(z)$), and is calculated based on $D(X)$, zero vector (i.e. $\mathbf{0}$) and a loss function (e.g. binary cross entropy). In each cycle, the weights of the discriminator network are updated based on the objective to minimize total loss (i.e. real loss + fake loss) and the weights of the generator will be updated based on its objective to maximize fake loss. This cycle continues until the stop condition (e.g. maximum number of epochs) is met. In an ideal training condition, both fake loss and real loss converges to 0.5 which indicates that it is impossible to distinguish between input real data and synthetic data because they are samples of the same distribution \cite{mirza2014cgan}. 

GAN and its variants have been widely used in different transportation applications recently including  real time crash detection \cite{cai2020real, islam2021crash, lin2020automated, chen2021traffic}, traffic flow data prediction  \cite{chen2019traffic, zhang2019trafficgan}, and autonomous driving \cite{kuefler2017imitating,arnelid2019recurrent} to name but a few. There are several GAN variants based on different architectures or loss functions proposed in the literature \cite{jabbar2021survey}. One of the popular variants is conditional-GAN (CGAN) \cite{mirza2014cgan} in which both the generator and the discriminator are conditioned on some data which could be a class label or a feature vector if we wish to use it for regression purposes. The goal of this paper is to propose a crash data augmentation method based on CGAN which can improve the performance of SPFs. The details of the method are presented in the next section.

\section{Methodology}
\label{section:Methodology}
Conditional generative adversarial network, or CGAN for short, is a strong deep generative model that has seen considerable applicability in many areas in recent years. To the best of the authors knowledge this is the first time CGAN is being used for crash frequency data augmentation for improving safety performance functions. The proposed method has two main steps. The first is to design and train CGAN based on the original crash data set,  and then use the trained generator to generate synthesized crash data to develop the SPF.

\subsection{Design and train CGAN}
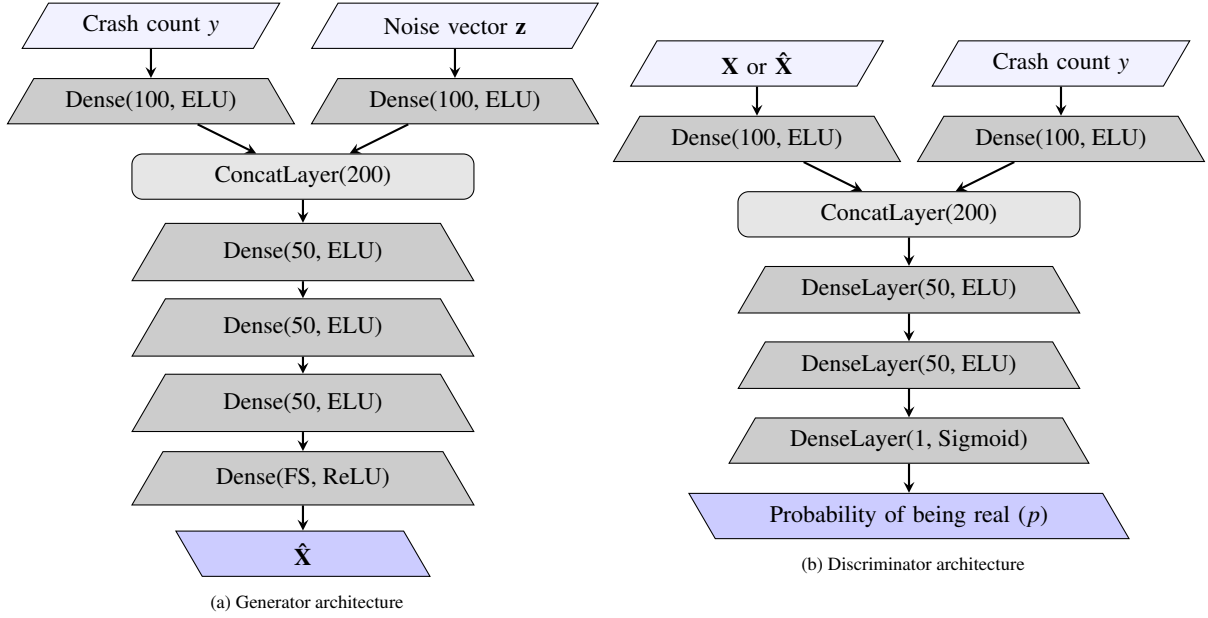
\begin{figure}
\small
  \centering
  \begin{subfigure}{0.48\textwidth}
    \begin{tikzpicture}[auto, node distance=4cm,>=latex']
        
        \node (concat) [process, minimum height = 0.6cm] {ConcatLayer(200)};
        \node (X_out) [NN, above of = concat, xshift=-2cm, yshift=-3cm, minimum height = 0.6cm, minimum width = 2cm] {Dense(100, ELU)};
        \node (X) [input, minimum height = 0.6cm,above of = X_out, xshift=0cm, yshift=-3cm] {Crash count $y$};
        
        \node (noise_out) [NN, above of = concat, xshift=2cm, yshift=-3cm, minimum height = 0.6cm, minimum width = 2cm] {Dense(100, ELU)};
        \node (noise) [input, minimum height = 0.6cm, above of = noise_out, xshift=0cm, yshift=-3cm] {Noise vector $\mathbf{z}$};
        
        \node (l1) [NN, below of = concat, xshift=0cm, yshift=3cm, minimum height = 0.6cm] {Dense(50, ELU)};
        \node (l2) [NN, below of = l1, xshift=0cm, yshift=3cm, minimum height = 0.6cm] {Dense(50, ELU)};
        \node (l3) [NN, below of = l2, xshift=0cm, yshift=3cm, minimum height = 0.6cm] {Dense(50, ELU)};
        \node (l4) [NN, below of = l3, xshift=0cm, yshift=3cm, minimum height = 0.6cm] {Dense(FS, ReLU)};
        
        \node (Gen_out) [output, minimum height = 0.6cm,below of = l4, xshift=0cm, yshift=3cm, text width=2cm] {$\mathbf{\hat{X}}$};

        \draw [arrow] (X) -- (X_out);
        \draw [arrow] (noise) -- (noise_out);
        \draw [arrow] (X_out) -- (concat);
        \draw [arrow] (noise_out) -- (concat);
        \draw [arrow] (concat) -- (l1);
        \draw [arrow] (l1) -- (l2);
        \draw [arrow] (l2) -- (l3);
        \draw [arrow] (l3) -- (l4);
        \draw [arrow] (l4) -- (Gen_out);

    \end{tikzpicture}
    
    \caption{Generator architecture}
    \label{fig:Generator architecture}
  \end{subfigure}
  \begin{subfigure}{0.48\textwidth}
    \begin{tikzpicture}[auto, node distance=4cm,>=latex']
        
        \node (concat) [process, minimum height = 0.6cm] {ConcatLayer(200)};
        \node (X_out) [NN, above of = concat, xshift=-2cm, yshift=-3cm, minimum height = 0.6cm, minimum width = 2cm] {Dense(100, ELU)};
        \node (X) [input, minimum height = 0.6cm, above of = X_out, xshift=0cm, yshift=-3cm] {$\mathbf{X}$ or $\mathbf{\hat{X}}$};
        
        \node (y_out) [NN, above of = concat, xshift=2cm, yshift=-3cm, minimum height = 0.6cm, minimum width = 2cm] {Dense(100, ELU)};
        \node (y) [input, minimum height = 0.6cm, above of = noise_out, xshift=0cm, yshift=-3cm] {Crash count $y$};
        
        \node (l1) [NN, below of = concat, xshift=0cm, yshift=3cm, minimum height = 0.6cm] {DenseLayer(50, ELU)};
        \node (l2) [NN, below of = l1, xshift=0cm, yshift=3cm, minimum height = 0.6cm] {DenseLayer(50, ELU)};
        \node (l3) [NN, below of = l2, xshift=0cm, yshift=3cm, minimum height = 0.6cm] {DenseLayer(1, Sigmoid)};
        
        \node (Dis_out) [output, minimum height = 0.6cm,below of = l3, xshift=0cm, yshift=3cm,  text width = 4cm] {Probability of being real ($p$)};

        \draw [arrow] (X) -- (X_out);
        \draw [arrow] (y) -- (y_out);
        \draw [arrow] (X_out) -- (concat);
        \draw [arrow] (y_out) -- (concat);
        \draw [arrow] (concat) -- (l1);
        \draw [arrow] (l1) -- (l2);
        \draw [arrow] (l2) -- (l3);
        \draw [arrow] (l3) -- (Dis_out);

    \end{tikzpicture}
    
    \caption{Discriminator architecture}
    \label{fig:Discriminator architecture}
  \end{subfigure}

\caption{Network architectures. The input layer of generator includes crash count (i.e. $y$) and a noise vector ($\mathbf{z}$) with the same size of the feature vector ($FS$), and input layer of discriminator includes normalized feature vector and crash count (i.e. $y$)} 
\label{fig:architecture}
\end{figure}

For crash frequency data augmentation, we used the generator and discriminator models shown in Figure \ref{fig:architecture}. \textit{Dense(n)} in Figure \ref{fig:architecture} is a regular deeply connected neural network layer with $n$ nodes and \textit{ConcatLayer(n)} concatenates a list of inputs. The first layer of each network is an input layer. For the generator, the input layer includes a noise vector ($z$) with the same size as the feature vector ($X$) and crash count ($y$) and the output is a generated feature vector ($\hat{X}$) with size of $FS$ (feature vector size). For the discriminator, crash counts ($y$), feature vector $X$ as real feature vector and $\hat{X}$ as generated feature vector are included in the input layer. The output of the discriminator is a number between 0 and 1 indicating the probability of being a real sample for the the given input. Also the activation functions used in the networks include Exponential Linear Unit (ELU), Rectified Linear Unit (ReLU), and Sigmoid \cite{sharma2017activation}.

After randomly initializing the weights of each network, in each cycle of CGAN training, the weights of the generator and discriminator networks are updated to minimize the corresponding loss functions \cite{mirza2014cgan}:

\begin{equation}
    Loss(D) = -\frac{1}{2} \left( \log(D(X|y)) + \log[1-D(\hat{X}|y)] \right)
\end{equation}

\begin{equation}
    Loss(G) = - \log(D(\hat{X}|y))
\end{equation}

\noindent where $D(X|y)$ and $D(\hat{X}|y)$ are the discriminator outputs given feature vector $X$ and generated feature vector $\hat{X}$ conditioned on crash count $y$. In an ideal equilibrium, both loss values will be equal to $-\log(0.5)$. In this case, the discriminator cannot distinguish between real and fake samples generated by the generator.

\subsection{Generate synthesized crash data and develop SPFs}
The generator network can be used to generate synthesized crash data once the CGAN has been trained. A random sample from the empirical distribution of crash counts and a random noise vector are input into the generator network for each synthesised data point. This procedure is continued until the desired number of fake data is attained. Finally, the safety performance functions are  developed using the augmented crash data set (i.e. real plus synthesized data). SPF development can be based on any approach accessible, including machine learning and traditional parametric models. In this study, we have used the NB model as the most common crash frequency modeling approach. The proposed method is implemented and evaluated using both simulated and real-world crash data sets. In the next section the results of these experiments are discussed.

\section{Experiments and Results}
\label{section:eval}
\subsection{Simulated Crash Data}
The performance of the proposed crash data augmentation method is investigated in this section in a simulated environment. Simulation enables us to perform the evaluation between base-SPF (the SPFs developed on the original data) and augmented-SPF (the SPFs developed on the augmented data) under various scenarios while also knowing the truth. The simulated data sets are created using the following three steps, which is consistent with the approach taken by previous studies  \cite{francis2012characterizing, zou2015modeling, ye2018semi}:

\begin{enumerate}
    \item Generate a random feature vector ($X$) with the size of 4 ($X_1, X_2, X_3, X_4$) from a uniform distribution on $[0, 1]$.
    \item Generate the corresponding crash count $Y$ from a Poison distribution with the long-term crash mean of $\lambda$ that is gamma distributed with the dispersion of $\alpha$:
    \begin{itemize}
        \item[] $Y \sim Poisson (\lambda)$;
        \item[] $\lambda = \exp(\beta_0 + 0.5X_1 - 0.5X_2 + X_3 - X_4 + \epsilon)$;
        \item[] $\exp(\epsilon) \sim Gamma (1,\alpha)$.
    \end{itemize}
    \item Steps (1) and (2) are repeated until the desired sample size is reached.
\end{enumerate}

\begin{figure}[tbp]
    \centering
    
    \begin{tikzpicture}[auto, node distance=4cm,>=latex']
        \node (train) [data, text width = 3cm] {\footnotesize Train data};
        \node (note1) [text width = 4cm, above of = train, yshift = -3cm, text centered] {\footnotesize One data set with the size of 100 for training CGAN};
         \node (test1) [data, right of = train, yshift = 1cm, xshift = 1cm, text width = 3cm] {\footnotesize NS test data 1};
         \node (test2) [data, below of = test1, yshift = 3.3cm, text width = 3cm] {\footnotesize NS test data 2};
         \node (testn) [data, below of = test2, yshift = 3.3cm, text width = 3cm] {...};
         \node (test1000) [data, below of = testn, yshift = 3.3cm, text width = 3cm] {\footnotesize NS test data 1000};
         \node (note2) [text width = 4cm, above of = test1, yshift = -3cm, text centered] {\footnotesize 1000 data sets with the size of 100 for SPF development};
         \node (note3) [text width = 4cm, below of = test1000, yshift = 3cm, text centered] {\footnotesize Performance measures: FI, PMD, MAPE (EB), MAPE (Dispersion)};

         \node (testa) [data, right of = test1, xshift = 1cm, text width = 4cm] {\footnotesize Prediction test data 1};
         \node (testb) [data, below of = testa, yshift = 3.3cm, text width = 4cm] {\footnotesize Prediction test data 2};
         \node (testc) [data, below of = testb, yshift = 3.3cm, text width = 4cm] {...};
         \node (testd) [data, below of = testc, yshift = 3.3cm, text width = 4cm] {\footnotesize Prediction test data 1000};
         \node (note4) [text width = 4cm, above of = testa, yshift = -3cm, text centered] {\footnotesize 1000 data sets with the size of 100 for SPF prediction testing};
         \node (note5) [text width = 5cm, below of = testd, yshift = 3cm, text centered] {\footnotesize Performance measure: MAPE (Crash frequency prediction)};

    \end{tikzpicture}
    
    \caption{Simulated data sets}
    \label{fig:data}
\end{figure}
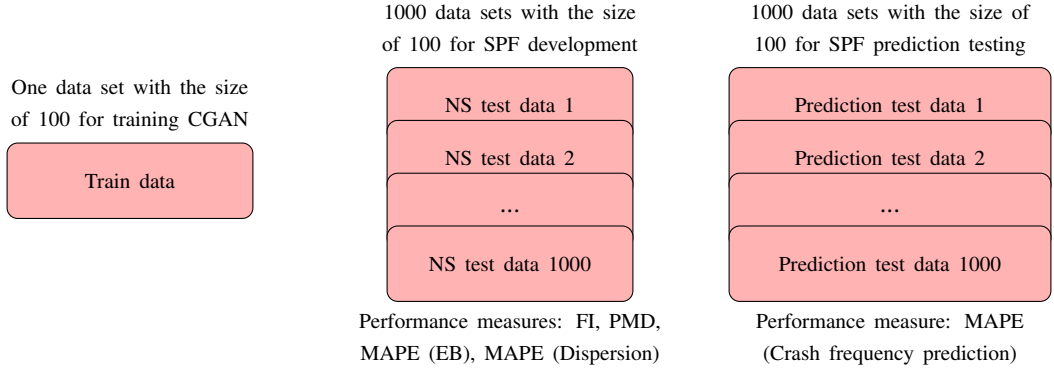

It is worth noting that $\epsilon$ indicates the unobserved heterogeneity that follows the log-gamma distribution, as specified in the NB model. The sample mean is determined by $\beta_0$.  As done by other studies \cite{francis2012characterizing, zou2015modeling, ye2018semi}  we set $\beta_0=0.5$ to represent a crash data set with low sample mean of 1.6.  Following the approach from these same studies, we evaluate two values of the dispersion parameter ($\alpha=0.5$ and $\alpha=1.5$) to represent low and high dispersed crash data sets.  The sample size for this experiment is set to 100 which is about 6 times smaller than the minimum sample size recommended by Lord et al (2006) for a sample mean of 1.6. To avoid over-fitting issues, CGAN was first trained on one simulated train data set (Figure \ref{fig:data}) with the size of 100 and then evaluated with a separate test data set. The training configuration parameters are set as follows:

\begin{itemize} [itemsep=0pt,parsep=0pt, topsep=0pt, partopsep=0pt]
    \item Optimizer: Adam \cite{pedamonti2018comparison}
    \item Number of epochs: 5000
    \item Batch size: 100
    \item Learning rate (both generator and discriminator): 0.001
    \item Learning rate decay (generator): 0.001
    \item Learning rate decay (discriminator): 0.0
\end{itemize}

\begin{figure}[]
  \centering
  \begin{subfigure}{0.49\textwidth}
    \includegraphics[width=\linewidth]{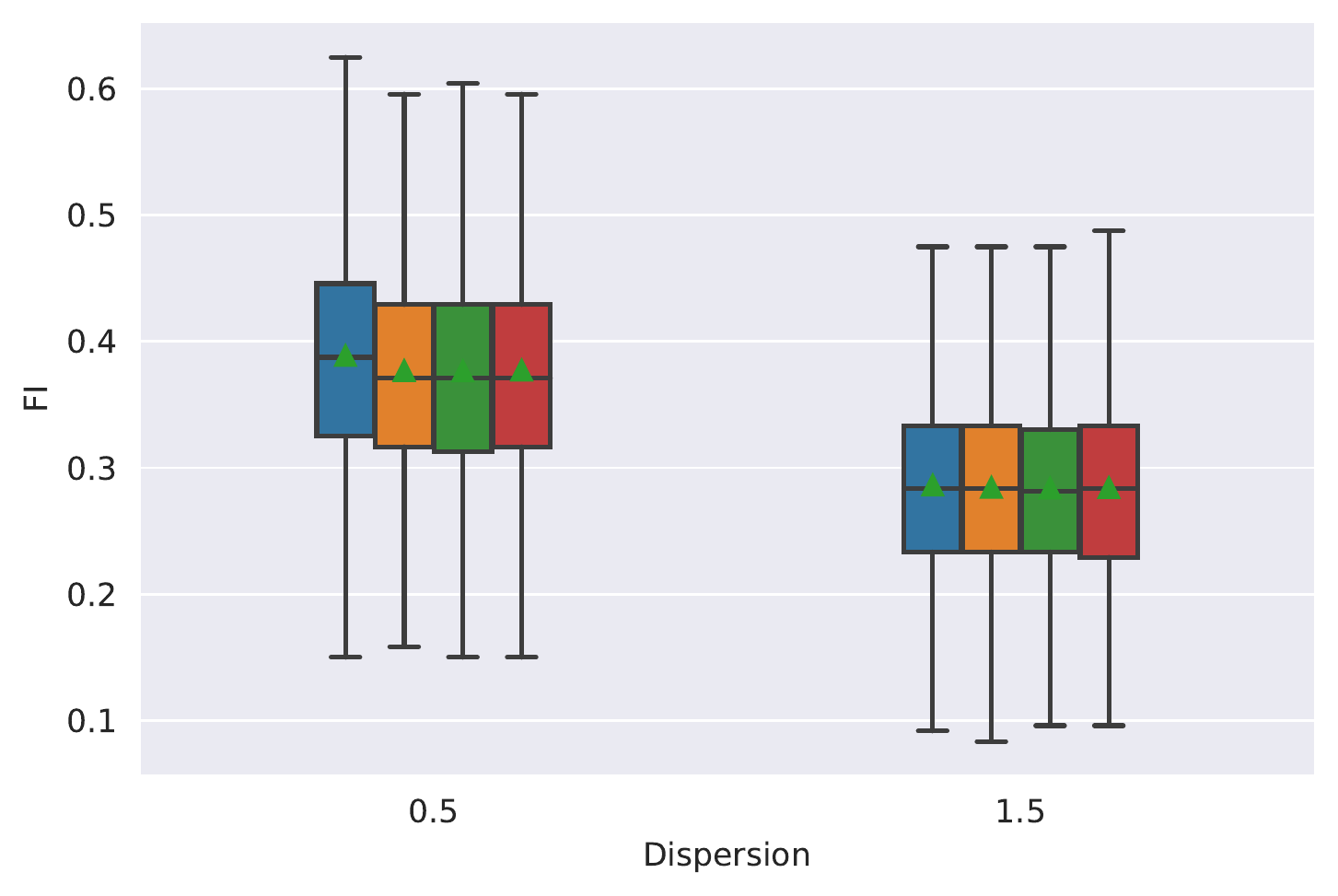}
    \caption{FI results} \label{fig:FI}
  \end{subfigure}
  \begin{subfigure}{0.49\textwidth}
    \includegraphics[width=\linewidth]{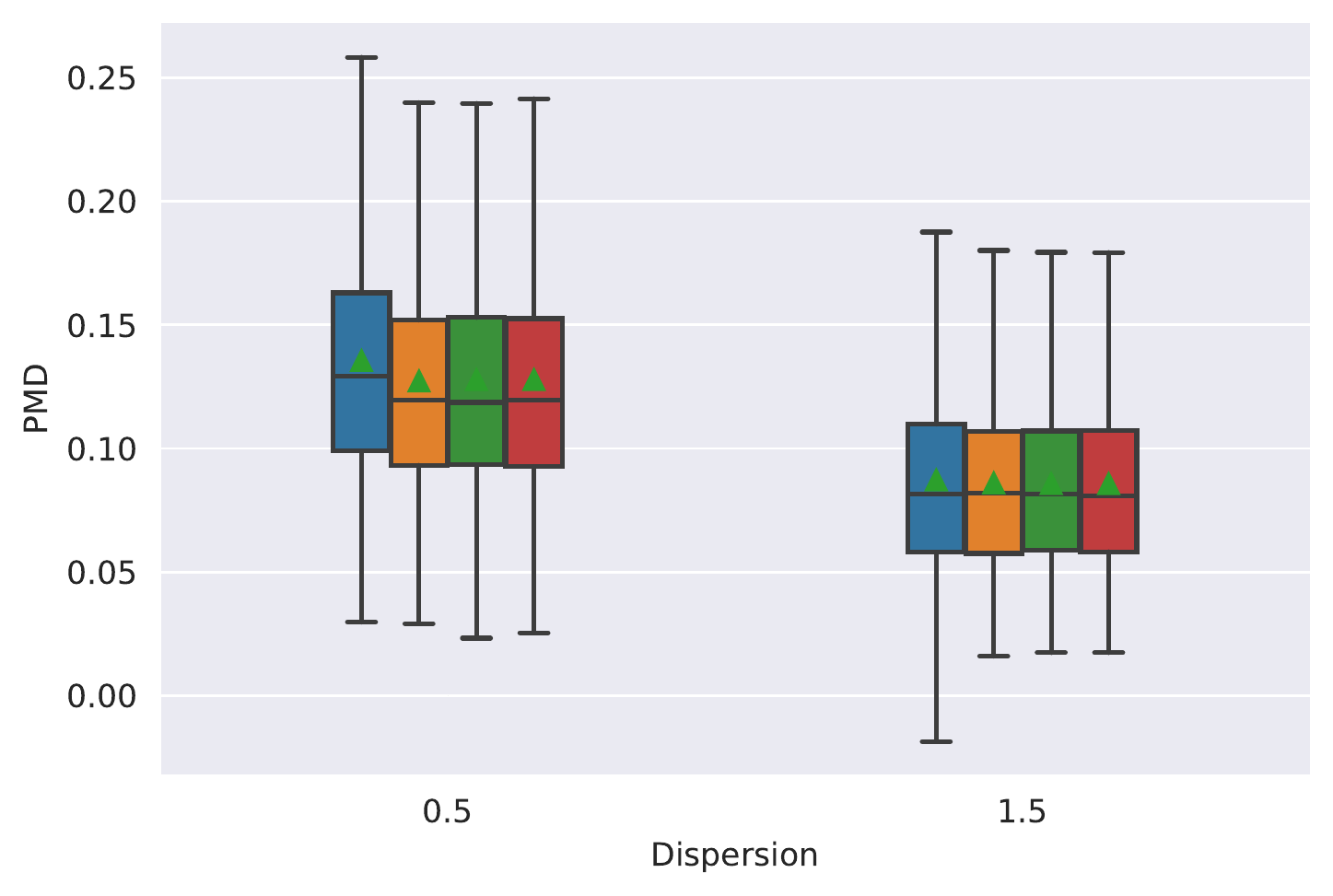}
    \caption{PMD results} \label{fig:PMD}
  \end{subfigure} \\
  
  \begin{subfigure}{0.49\textwidth}
    \includegraphics[width=\linewidth]{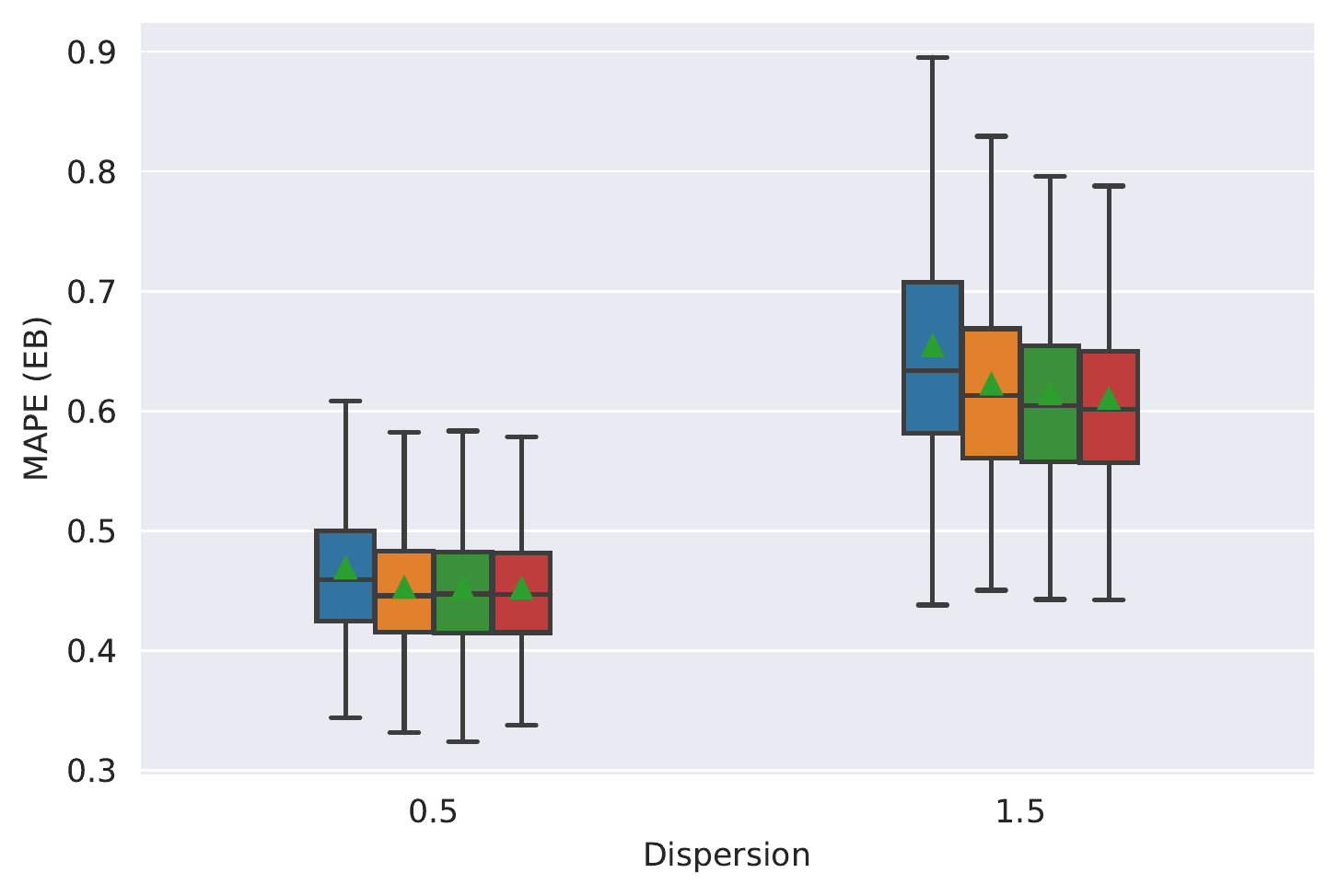}
    \caption{MAPE for EB estimates} \label{fig:EB}
  \end{subfigure}
  \begin{subfigure}{0.49\textwidth}
    \includegraphics[width=\linewidth]{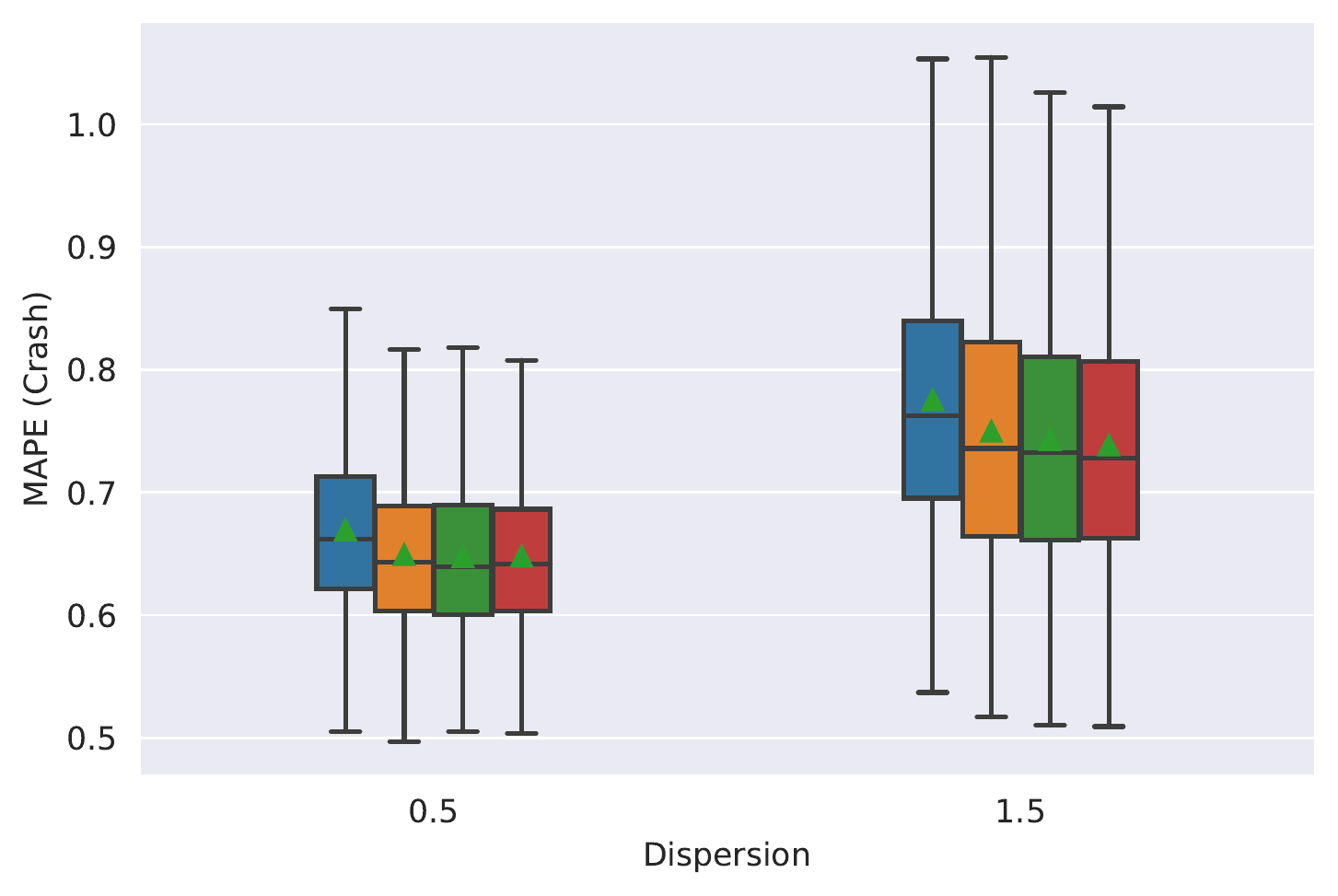}
    \caption{MAPE for crash frequency estimates} \label{fig:Crash}
  \end{subfigure} \\
  \begin{subfigure}{0.6\textwidth}
    \includegraphics[width=\linewidth]{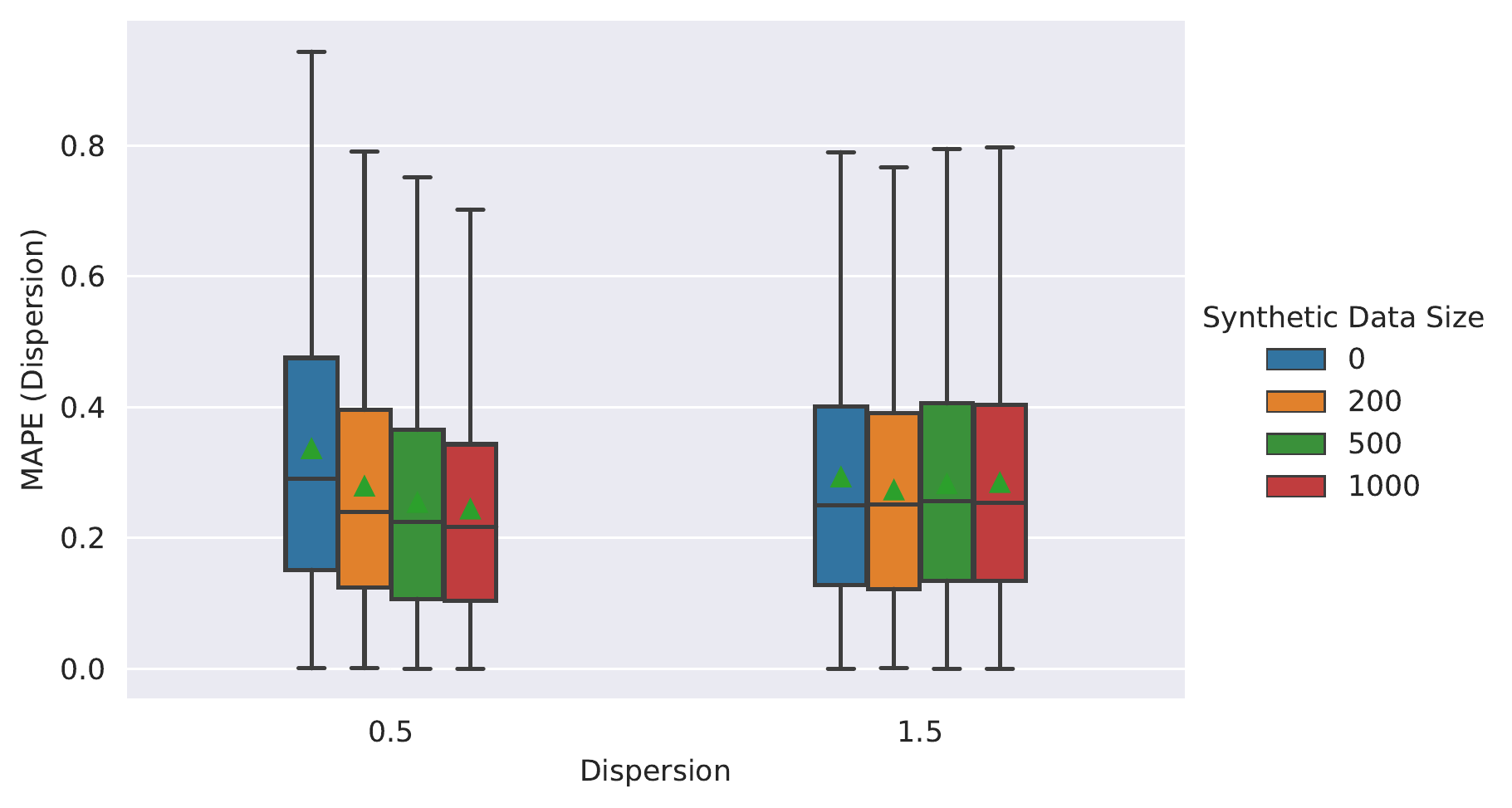}
    \caption{MAPE for dispersion parameter estimates} \label{fig:dispersion}
  \end{subfigure}

\caption{Simulation results} 
\label{fig:results}
\end{figure}

The evaluation was conducted using 1000 different simulated test data sets (NS test data in Figure \ref{fig:data}). Base-SPFs are developed using simulated test data sets (each set consists of 100 simulated observations) and augmented-SPFs are developed using the augmented test data sets (which contain the simulated test data and the synthesized data generated by the trained CGAN model). The two sets of SPFs are compared in terms of hotspot identification performance, model prediction accuracy, and dispersion parameter estimation accuracy. The results are summarized in Figure \ref{fig:results} and Table \ref{table:results}, and discussed in the following sections.

\begin{table}[ht]
\centering
\caption{Average performance improvements (\%) for Augmented-SPFs versus Base-SPFs}
\begin{tabular}{ccccccc}
\toprule
                                       \textbf{Dispersion} & \textbf{Synthetic   Data Size} & \textbf{FI}    & \textbf{PMD}   &\textbf{ MAPE (Crash)} & \textbf{MAPE (EB)} & \textbf{MAPE (Dispersion)} \\ \midrule
\multirow{3}{*}{0.5} & 200               & 3.1\% & 6.1\% & 3.0\%     & 3.5\%   & 17.1\%     \\
                                       & 500               & 3.2\% & 6.0\% & 3.4\%     & 3.7\%   & 29.5\%     \\
                                       & 1000              & 3.1\% & 6.0\% & 3.3\%     & 3.7\%   & 36.4\%        \\ \midrule
\multirow{3}{*}{1.5} & 200               & 0.7\% & 1.4\% & 3.3\%     & 4.9\%   & 6.8\%      \\
                                       & 500               & 0.9\% & 1.7\% & 4.3\%     & 6.4\%   & 3.9\%      \\
                                       & 1000              & 0.8\% & 1.7\% & 5.0\%     & 7.2\%   & 3.1\%     \\ \bottomrule

\end{tabular}
\label{table:results}
\end{table}

\subsubsection{Hotspot identification performance}
For each test data set, the hotspot identification process is performed using both base-SPFs and augmented-SPFs. The process includes ranking sites based on the EB estimates (i.e. estimation of long-term crash mean $\lambda$) \cite{hauer1997observational}:

\begin{equation}
    \label{Eq:EB}
    EB= \frac{1}{1+\alpha \mu} \times \mu + \frac{\alpha\mu}{1+\alpha\mu}\times y 
\end{equation}
\noindent where $\mu$ is the crash frequency prediction from either base-SPFs or augmented-SPFs. 
The hotspots suggested based on each SPF can be compared to the true hotspots (ranked based on $\lambda$) using False Identification (FI) and Poisson Mean Difference (PMD) tests \cite{cheng2008test}. The FI test shows the percentage of sites that are erroneously categorised as hotspots, and the PMD test is the relative difference of the sum of the Poisson means ($\lambda$) for true hotspots and the suggested hotspots based on EB estimates. FI and PMD are computed based on a given number of top hotspots. In this experiment, we calculate FI and PMD for the top 5, 10, 15 and 20 hotspots and then calculate the average of the four values. The FI and PMD results for this experiment for 1000 different test data sets are presented in Figure \ref{fig:FI} and Figure \ref{fig:PMD} as box-plots (box depicts minimum, first quartile (Q1), median, third quartile (Q3), and maximum, and the solid green triangles represent the mean).

The results for a dispersion value of 0.5 show that augmentation reduced the average of FI and PMD (green triangles in Figure \ref{fig:FI} and Figure \ref{fig:PMD}) by about 3\% and 6\%, respectively (see Table \ref{table:results}. The improvement was less substantial for the dispersion of 1.5, around 1\% for FI and 2\% for PMD. This difference was expected because, according to Equation \ref{Eq:EB}, SPF predictions have a greater influence on EB estimate when dispersion is lower. Consequently, the ranking cannot be dramatically changed by improving SPFs.

\subsubsection{Accuracy of model predictions}
The prediction performance of Base and Augmented SPFs are compared in terms of mean absolute percentage error (MAPE) for EB estimates. In addition, the MAPE of the SPF predictions for crash frequencies of 1000 crash frequency prediction test data  (Figure \ref{fig:data}) are calculated. The results are presented in Figure \ref{fig:EB} and Figure \ref{fig:Crash}.

Based on these results, there is significant improvement in the accuracy of EB estimation accuracy and crash frequency prediction. The improvement for EB estimates are about 4\% for low dispersed data sets, and about 5-7\% for high dispersed data sets. The reduction in MAPE of crash frequency predictions is about  3.5\% and 4.5\% for the dispersion of 0.5 and 1.5 respectively (refer to Table \ref{table:results}).

\subsubsection{Accuracy of dispersion estimation}
The dispersion parameter for each SPF for each test data set is determined using auxiliary Ordinary Least Square (OLS) regression without constant \cite{aux_OLS}. The MAPE for dispersion estimation for original data set and augmented data sets are also calculated and presented in Figure \ref{fig:dispersion}. The results show that crash data augmentation improved the dispersion parameter estimation accuracy up to about 36\% for low dispersion data sets but only around 4\% for high dispersion data sets.

The simulation results confirmed that the Augmented-SPFs outperform the Base-SPFs in almost all performance measures. The t-test results confirmed that these improvements are statistically significant. Regarding the size of synthetic data set, the t-test results indicate the difference between improvement caused by increasing the size of synthetic data is statistically significant for MAPE of EB estimates (dispersion = 1.5), MAPE of crash frequency prediction (dispersion = 1.5), and MAPE for dispersion parameter estimation (dispersion = 0.5). In the other cases, the improvement was not statistically significant. In the next section, the performance of Augmented-SPFs is evaluated using a real-world crash data set. 

\subsection{Real-world crash data}
A real-world crash data set is also used to assess the performance of the proposed crash data augmentation method. The data include crash counts and traffic volumes of minor and major streets of 200 4-legged intersections with stop sign control located in London, ON, Canada for 2017. The data set is randomly split into a train and a test data set with the ratio of 1 to 1. CGAN is trained using the train set with the same configuration setup as used in the simulation experiment. The trained generator is then used to produce synthesized crash data set. 

\subsubsection{Feature distribution evaluation}
\begin{figure}[h]
    \centering
    \includegraphics[width=0.8\linewidth]{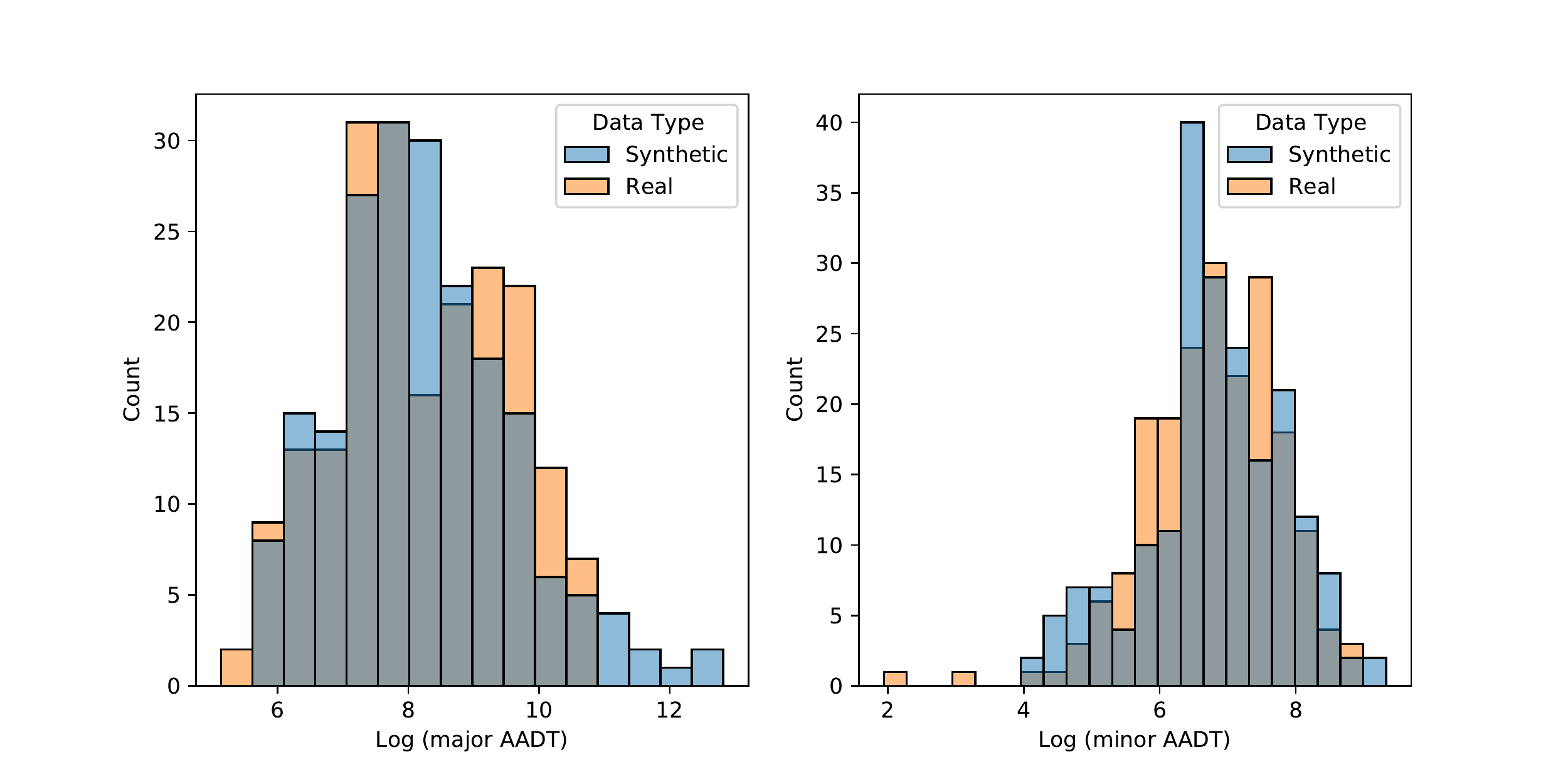}
    \caption{Distribution of major/minor AADT for synthetic data versus real data}
    \label{fig:dist}
\end{figure}

The distributions of two features (major and minor AADT) for real and synthesized crash data sets are shown in Figure \ref{fig:dist}. For both features, the distribution of the generated data closely matches the distribution of the real data. To confirm this observation statistically, three tests were carried out. The t-test shows that there is not evidence to conclude that the  mean of the two distributions are significantly different (p-value = 0.33 for major AADT and p-value = 0.36 for minor AADT). The Leven test shows that there is no evidence to conclude that the differences in the variance of the distributions are statistically significant (p-value of 0.47 for major AADT and 0.53 for minor AADT). Finally, the Kolmogrove-Smirnov test shows that there is no evidence to conclude that the distributions of synthesized data set and real data set are statistically different.  

\subsubsection{SPF evaluation}
Base SPFs developed using the train data set and Augmented SPFs developed using augmented  data set (with 1000 synthesized observations) are as follows:

\begin{equation}
    SPF_{Base} = \exp(-4.64 + 0.53 \times \ln(AADT_{major}))
\end{equation}

\begin{equation}
    SPF_{Augmented} = \exp(-5.69 + 0.42 \times \ln(AADT_{major}) + 0.20 \times \ln(AADT_{minor}))
\end{equation}

\noindent Minor AADT was not a statistically significant feature in the Base SPF (p-value = 0.28 $>$ 0.05) while it was a significant feature in the Augmented SPF. The prediction performance of these two SPFs are also compared in terms of the MAPE of crash frequency prediction using the test data set. The MAPE was 0.70 and 0.59 for the Base SPF and Augmented SPF respectively, indicating that the data augmentation provided approximately a 14\% improvement in crash frequency prediction accuracy.

\section{Conclusions and Recommendations}
\label{section:conclusion}
SPFs are commonly used in road network safety screening to identify and prioritize crash hotspots and to assess the efficiency of safety countermeasures. One of the challenges for developing reliable SPFs is dealing with crash data set with low sample sizes and low sample mean. Traditionally, the data set is expanded by using a few years of crash counts for each site. This method assumes that the network characteristics remain the same during the study period which is not always valid. 

In this paper, a data augmentation method for crash frequency data based on conditional generative adversarial networks (CGAN) is proposed, which can fully mimic the underlying distribution of the given crash data and generate synthesised samples. The generated samples can then be added to the original data set for SPF development purposes. Based on simulation results, False Identification (FI) and Poison Mean Difference (PMD) tests for Augmented SPFs on average were improved by about 1-3\% and 2-6\% respectively in comparing to the results for Base SPFs. Also, Augmented SPFs showed lower mean absolute percentage errors (MAPE) for EB estimates, crash frequency predictions, and dispersion parameter estimation (4-7\%, 3.5-4.5\%, 4-30\% reduction respectively). The results indicate that larger improvements are related to simulated data set with low dispersion parameter values. 

The method is also evaluated using a real-world crash data set for 200 stop-controlled intersections (with 50/50 train/test split) with minor and major traffic volumes. Statistical tests showed that the distribution of the generated sample is not significantly different from the original data set. Both features (minor and major AADT) were found significant in the Augmented SPF while minor AADT was not detected as a significant variable for Base SPF. In addition, the MAPE result for the test data set using Augmented SPFs showed about 14\% improvement.

Although the results are encouraging, there are several limitations in the current work which remains for future studies. For example, one type of traffic location (4-legged stop-controlled intersections) is used in this study to evaluate the effectiveness of the proposed method. It should be investigated if one single CGAN can model the distribution of a crash data set including different types of facilities. Further, there are several types of GAN models in terms of architecture and loss functions. Future studies can focus on finding the best type suited for crash data modeling.






\bibliographystyle{Other/model1-num-names.bst}
\bibliography{main.bib}







\end{document}